\documentclass{article}
\usepackage{ijcai23}
\usepackage{times}
\usepackage{soul}
\usepackage{url}
\usepackage[hidelinks]{hyperref}
\usepackage[utf8]{inputenc}
\usepackage[small]{caption}
\usepackage{graphicx, subfigure}
\usepackage{amsmath}
\usepackage{amsthm}
\usepackage{booktabs}
\usepackage{algorithm}
\usepackage{algorithmic}
\usepackage[switch]{lineno}
\linenumbers

\begin{document}
\nolinenumbers

\title{Analysis on Effects of Fault Elements in Memristive Neuromorphic Systems}

\author{
Hyun-Jong Lee$^1$
\and
Jae-Han Lim$^1$\thanks{Corresponding author}\and
\affiliations
$^1$Department of Software \\ Kwangwoon University \\
\emails
{hetzer44, ljhar}@kw.ac.kr
}

\maketitle

\begin{abstract}
Nowadays, neuromorphic systems based on Spiking Neural Networks (SNNs) attract attentions of many researchers. There are many studies to improve performances of neuromorphic systems. These studies have been showing satisfactory results. To magnify performances of neuromorphic systems, developing actual neuromorphic systems is essential. For developing them, memristors play key role due to their useful characteristics. Although memristors are essential for actual neuromorphic systems, they are vulnerable to faults. However, there are few studies analyzing effects of fault elements in neuromorphic systems using memristors. To solve this problem, we analyze performance of a memristive neuromorphic system with fault elements changing fault ratios, types, and positions. We choose neurons and synapses to inject faults. We inject two types of faults to synapses: SA0 and SA1 faults. The fault synapses appear in random and important positions. Through our analysis, we discover the following four interesting points. First, memristive characteristics increase vulnerability of neuromorphic systems to fault elements. Second, fault neuron ratios reducing performance sharply exist. Third, performance degradation by fault synapses depends on fault types. Finally, SA1 fault synapses improve performance when they appear in important positions.
\end{abstract}

\section{Introduction}
Dramatic increase of data in modern society has triggered neural networks to grow rapidly. Various neural networks currently appear. They show excellent performances in many fields. Therefore, researchers try to make computer systems that are affable to neural networks. In one of these attempts, researchers focus on neuromorphic systems based on Spiking Neural Networks (SNNs). Neuromorphic systems mimic human brains to gain the two important advantages from the brains: event-driven operations and parallel structures. These advantages prevent them from calculating weights too frequently and using resources for large processors and memories to communicate with each other. 

Because researchers have developed useful learning mechanisms for neuromorphic systems, neuromorphic systems show impressive performances in various fields: classification, recognition, and prediction \cite{result1} \cite{result2}. However, there are few studies explaining how neuromorphic systems operate in actual devices although neuromorphic systems maximize their advantages with the actual devices. A lack of researches dealing with actual neuromorphic systems makes it difficult to use the advantages of neuromorphic systems completely. To build actual neuromorphic systems, memristors are important components that are used most widely \cite{reconfig}. Memristors have the two following characteristics. First, they can represent synaptic weights as resistance. Second, they don't need power to maintain memories. Due to these characteristics, they are suitable to create actual neuromorphic systems.

Although memristors are necessary for developing actual neuromorphic systems, memristors are vulnerable to fault elements \cite{defect}. There are previous studies that analyze effects of fault elements in neuromorphic systems \cite{guidance}. However, conventional studies analyze them without memristive characteristics \cite{fault} \cite{avoid}. Analyzing the effects without memristive characteristics cannot provide guidelines to deal with fault elements in actual neuromorphic systems. Therefore, it is essential to analyze the effects of fault elements with memristive characteristics.

To address this problem, we analyze various effects of fault elements using a neuromorphic system simulator that has memristive characteristics. We create fault neurons that do not emit output spikes at all. For creating fault synapses, we use two types of fault synapses: Stuck-at-0 (SA0) fault synapses and Stuck-at-1 (SA1) fault synapses. SA0 fault synapses cannot deliver spikes to post-synaptic neurons and SA1 fault synapses pass spikes to post-synaptic neurons without resistance. We measure performance changes caused by fault neurons and synapses with various scenarios. The contributions of this paper as follows.

\begin{flushleft}
$\bullet$ To our knowledge, this is the first in-depth work to analyze effects of fault elements in a memristive neuromorphic system.
\\

$\bullet$ Our work shows performance changes depending on fault synapse types and fault synapse positions corresponding to data features. 
\\

$\bullet$ We find interesting effects of fault elements. First, neuromorphic systems require specific number of neurons that store data features to ensure performances according to the number of data classes. Second, SA1 fault synapses can improve performances when they appear in important positions corresponding to significant features of input data.

\end{flushleft}

The remainder of this paper is organized as follows. In Section 2, we introduce the latest technological trends and fault element researches in neuromorphic systems. In Section 3, we briefly explain SNNs, neuromorphic systems, memristors, and fault elements. In Section 4, we explain our system’s mechanisms and fault modeling methods. In Section 5, we analyze effects of fault elements with various scenarios. This paper concluded in Section 6.
\section{Related Works}
\subsection{Applications and developments of neuromorphic systems}
Studies dealing with neuromorphic systems have applied them to various tasks. They have shown good performance in image identification with low energy consumption \cite{consumption}. Furthermore, they can reduce complexity of detection models in sequential data while maintaining high performance \cite{reduction}. From the point that neuromorphic systems are suitable to analyze sequential data, researchers try to use them for interpreting biological signals that are one-dimensional. They can classify Electroencephalography (EEG) with high performance \cite{eeg}. They have more efficient learning rule for Electrocardiogram (ECG) signals and show very low energy consumption \cite{wearable}.

Due to the rising popularity of neuromorphic systems, lots of studies have developed useful mechanisms for them. Many studies focus on developing new learning rules and implementation methods. Through event based computing and reward modulated learning rule, neuromorphic systems are more energy efficient than the previous systems \cite{reward}. Assembly-based Spike Timing Dependent Plasticity (STDP) helps them to extend layers with performance gain \cite{assembly}. Back-Propagation (BP) methods are efficient learning rules in neuromorphic systems. Memristors allow neuromorphic systems to use neural networks with BP and they have the potential to apply neuromorphic systems to intelligent robot system \cite{reconfig}.  

\subsection{Analysis and solutions for fault elements in neuromorphic systems}
Due to fault vulnerabilities of neuromorphic systems, researchers have analyzed how fault elements affect neuromorphic systems. The faults usually occur due to complexity of neuromorphic systems and they reduce performances. They find the bad effects of fault neurons and synapses in many cases \cite{guidance}. With the increasing interests in memristors, researchers analyze how faults occur in memristors \cite{defect}. However, they have not analyzed how faults affect memristive neuromorphic systems in various situations.

Through studying characteristics of fault elements, researchers develop mechanisms to increase fault tolerance. They regulate network structure and lead neuromorphic systems to avoid sending spikes to fault neurons \cite{avoid}. In addition, they propose a fault tolerant neuromorphic system with self repairing capability in network topology \cite{repair}. Researchers do not only create fault tolerant network but they also create a fault tolerant learning rule. They develop a fault resistant algorithm based on fault aware mapping and training and this algorithm improves the accuracy by up to 70\% compared to a baseline \cite{respawn}.
\section{Backgrounds}
\subsection{Spiking Neural Networks (SNNs) and Neuromorphic systems}
SNNs are neural networks using spikes to process data. In SNNs, there are spiking neurons having membrane potential. A spiking neuron increases their membrane potential whenever they receive membrane potential. If the potential reaches a threshold, the neurons emit spikes. This operation makes SNNs have an event-driven operation and leads SNNs to be more energy efficient than other neural networks. In neuromorphic systems, there are many neurons and synapses in parallel. Different from Von Neumann systems, these neurons and synapses do not need to communicate with each other because they synchronize each other simultaneously. This point prevents neuromorphic systems from consuming energy for the communication and makes the them use less energy than Von Neumann systems. 

SNNs and neuromorphic systems are essential to each others. SNNs can show higher performances using neurons and synapses in neuromorphic systems. Neuromorphic systems can save more energy using the event-driven operation of SNNs. SNNs and neuromorphic systems are complementary to each other. Thus, most neuromorphic systems are based on SNNs and SNNs operate in neuromorphic systems.

\subsection{Memristor}
Memristors are key elements for developing actual neuromorphic systems. They are tiny passive elements act as conductors and resistors. In memristors, conductance and resistance are synchronized with each other. When the conductance increases, the resistance decreases simultaneously. Therefore, memristive neuromorphic systems can detect synaptic weight changes easily. Furthermore, memristors do not always need power source to remember weights. Although memristors have many advantages for neuromorphic systems, they update synaptic weights unstably since they change their conductance non-linearly \cite{nonlinear}. Furthermore, they break down easily because of their complicated structures \cite{defect}. 

\subsection{Fault elements}
In neuromorphic systems, fault neurons and synapses often appear. They reduce performances of neuromorphic systems severely. Specifically, fault neurons do not generate output spikes and fault synapses don't change their weights. The fault synapses that frequently appear in neuromorphic devices are Stuck-At-0 (SA0) fault synapses and Stuck-At-1 (SA1) fault synapses. SA0 fault synapses don't pass input spikes to post-synaptic neurons at all because their resistance is fixed to infinite. They prevent post-synaptic neurons from increasing membrane potential and the neurons cannot learn data samples properly. SA1 fault synapses receive input spikes without any resistance because their resistance is fixed to the minimum value. They cause post-synaptic neurons to increase membrane potential easily and the neurons emit spikes too frequently. Therefore, SA0 and SA1 fault synapses cause severe performance degradation in neuromorphic systems. 
\section{Explanations on the neuromorphic system}
\subsection{Neuromorphic system structure}
\begin{figure}[htb!]
\centering
\includegraphics[width=250pt]{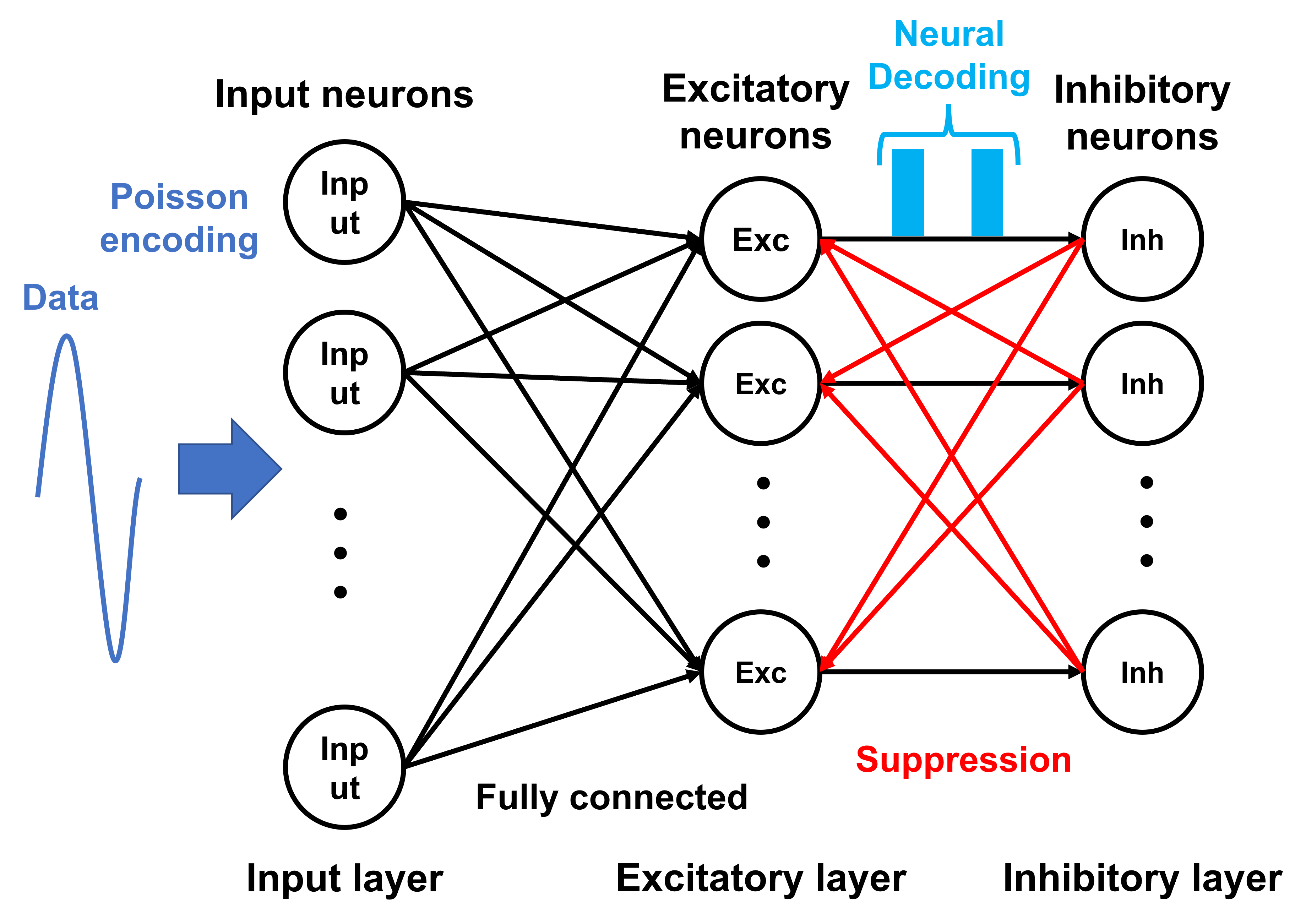}
\caption{The neural network consists of 3 layers: input layer for generating input spikes, excitatory layer for learning, and inhibitory layer for choosing proper markers.}
\label{fig:structure}
\end{figure}

We use Diehl\&Cook2015 structure to build a neural network because it is proved that this structure shows promising performances \cite{Diehl}. The neural network in our neuromorphic system consists of three layers: input layer, excitatory layer, and inhibitory layer. Figure \ref{fig:structure} shows overall structure of the neural network. The input layer consists of input neurons that generate input spikes from data using Poisson encoder. The excitatory layer consists of excitatory neurons that are Leaky Integrate-and-Fire (LIF) neurons with adaptive thresholds for maintaining homeostasis of neurons. They receive input spikes from input layer and emit output spikes to the inhibitory layer. The inhibitory layer consists of inhibitory neurons that are LIF neurons without adaptive thresholds. Inhibitory neurons emit suppression spikes to prevent unnecessary excitatory neurons from emitting spikes.

Excitatory neurons store features of data as synaptic weights and their synaptic weights change according to Spike Timing Dependent Plasticity (STDP). Inhibitory neurons suppress the excitatory neurons that do not learn data features appropriately. The suppressed neurons cannot emit output spikes and neurons that learn data features properly emit output spikes. Therefore, the system predicts answer correctly. To predict answers from output spikes, we use a neural decoder. The decoder assigns a marker indicating the class of data to excitatory neurons. After the decoder assigns markers, the system prints out a marker of the most frequently firing excitatory neuron as the classification result.

\subsection{Neuron modeling}
An LIF neuron model is the most widely used model in SNNs due to its simplicity. LIF neurons operate with the following sequences. First, when an LIF neuron receives input spikes, the spikes raise the membrane potential in proportion to synaptic weights. Second, if its potential reaches the threshold, the neuron fires and emits an output spike. We apply adaptive thresholds to excitatory neurons because adaptive thresholds help them to store features of input data \cite{theta}. The adaptive threshold factor named $\theta+$ makes thresholds of excitatory neurons rise whenever an excitatory neuron fires and it becomes hard for excitatory neurons to fire. Therefore, adaptive thresholds prevent excitatory neurons from firing too frequently and keeps them generating output spikes stably. The membrane potential of LIF neurons and adaptive threshold are expressed by following equations

\begin{equation}
\label{eq:membrane}
\tau_{m} \frac{dV_{m}(t)}{dt}=-V_{m}(t)+R\sum_{i=1}^{n_{l}}(w_{i}\sum_{k}\delta_{i}(t-t_{k}))
\end{equation}

\begin{equation}
\label{eq:threshold}
\tau_{\theta} \frac{d\theta(t)}{dt}=V_{thres}-\theta(t)+\tau_{\theta}\theta_{plus}\delta(t-t_{x})
\end{equation}
\\
where $t$ is current time, $V_{m}$ is a current voltage of membrane potential, and $\tau_{m}$ is a time constant of the membrane potential. $R$ is a resistance of a synapse between neurons, $n_{l}$ is the number of pre-synaptic neurons of the $l_{th}$ layer, and $w_{i}$ is the weight of the $i_{th}$ neuron in $l_{th}$ layer. $\delta_{i}(t-t_{k})$ is a spiking event from $i_{th}$ neuron and $\delta(t-t_{x})$ is a spiking event of post-synaptic neurons. $\tau_\theta$ is a time constant of the threshold $\theta$ and $V_{thres}$ indicates an initial voltage of the threshold. $\theta$ is current threshold and $\theta_{plus}$ is a constant value that increases the threshold whenever spiking events occur (i.e., $\delta(t-t_{x})$).

\subsection{Spike Timing Dependent Plasticity (STDP)}
STDP is the most focused learning rule due to its bio-plausible characteristic and fast learning speed \cite{fast}. STDP controls synaptic weights through Long Term Potentiation (LTP) and Long Term Depression (LTD). LTP increases synaptic weights and LTD decreases the weights. STDP determines LTP and LTD according to order of pre-synaptic and post-synaptic spikes. In our neuromorphic system, an STDP module controls synapses between input and excitatory neurons. 

\begin{figure}[htb!]
\centering
\includegraphics[width=225pt]{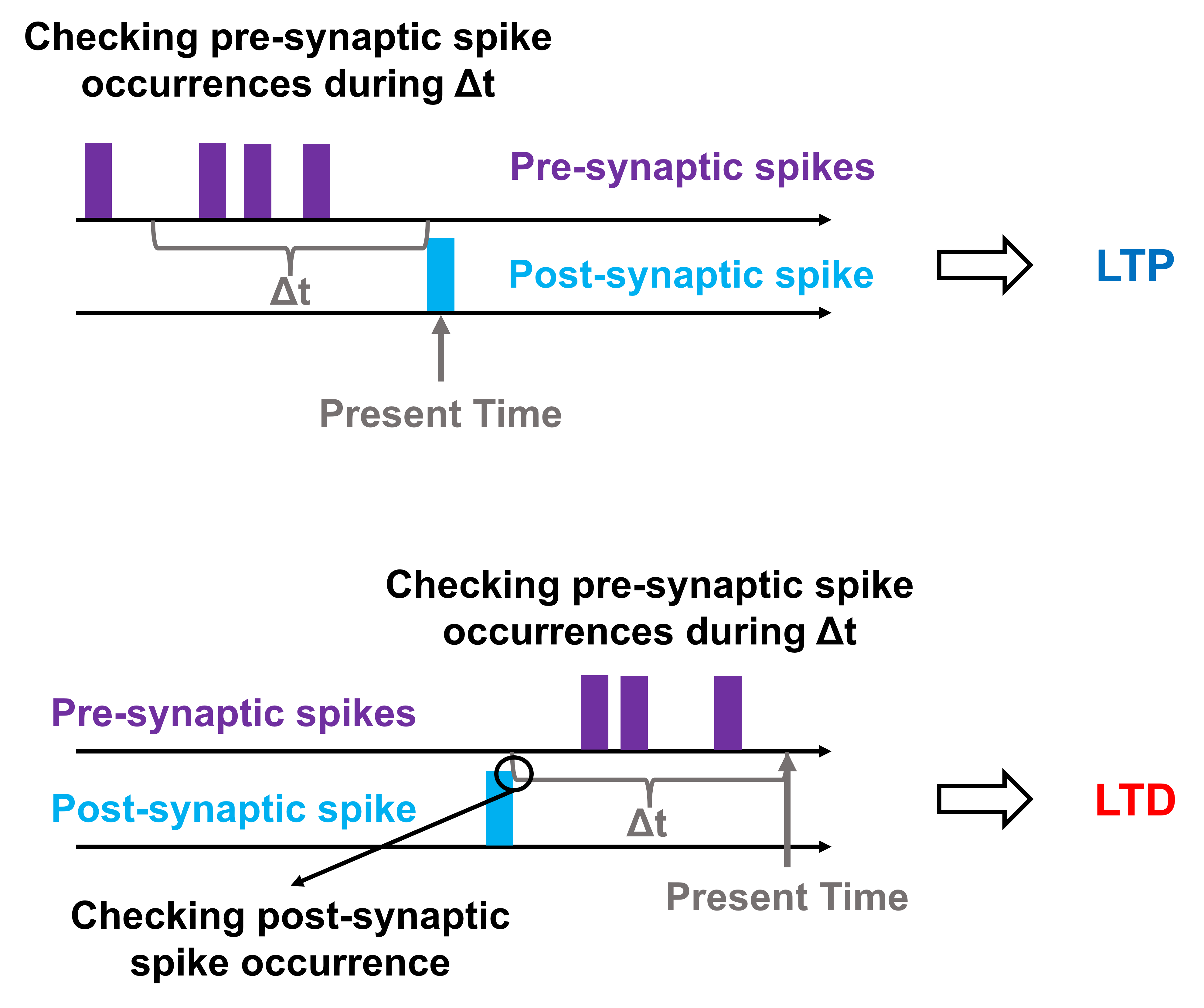}
\caption{Our STDP module decides LTP or LTD through spiking order of pre-synaptic neurons and post-synaptic neurons. When the module recognizes occurrence of post-synaptic spikes, it checks pre-synaptic spikes appear during $\Delta t$ and chooses between LTP or LTD.}
\label{fig:STDP}
\end{figure}

Figure \ref{fig:STDP} depicts how our STDP module decides LTP and LTD. When a post-synaptic neuron fire, STDP module checks whether pre-synaptic neurons connected to the post-synaptic neuron have fired during $\Delta t$. If they have fired, then STDP module determines LTP occurs and increase synaptic weights between the fired pre-synaptic neurons and the post-synaptic neuron in proportion to the number of the pre-synaptic spikes. STDP module also checks whether a post-synaptic neuron fired before $\Delta t$ from present time. If the post-synaptic neuron fired, STDP module checks whether pre-synaptic neurons connected to the post-synaptic neuron have fired during $\Delta t$. If they have fired, then STDP module determines LTD has occurred and decreases synaptic weights between the fired pre-synaptic neurons and the post-synaptic neuron in proportion to the number of the pre-synaptic spikes. 

To reflect memristive characteristics, STDP module changes synaptic weights non-linearly through the following equations \cite{resistive} \cite{nonlinear}

\begin{equation}
\label{eq:alpha}
\alpha=\frac{w_{max}-\ w_{min}}{1-\ e^{-v}}
\end{equation}

\begin{equation}
\label{eq:LTP}
\Delta w=\frac{\Delta t}{t_{post}-t_{pre}+1}(\alpha+w_{min}-w)(1-e^{(-\beta \frac{v_{LTP}} {256})})
\end{equation}

\begin{equation}
\label{eq:LTD}
\Delta w=-\frac{\Delta t}{t_{pre}-t_{post}+1}(\alpha-w_{max}+w)(1-e^{\frac{v_{LTD}} {256}})
\end{equation}
\\
where $w$ is the current synaptic weight and $\Delta w$ is the amount of the change in synaptic weights. $w_{min}$ and $w_{max}$ are the minimum and maximum value of the weights, respectively. We randomly set $w_{min}$ from 0 to 0.5 and $w_{max}$ from 0.5 to 1. $v$ is a parameter deciding the non-linearity and we divide $v$ into two cases: 1) $v_{LTP}$ for LTP and 2) $v_{LTD}$ for LTD. $\beta$ is for enabling different effect ratio between LTP and LTD. $\Delta t$ is the reference time for demonstrating LTP and LTD. $t_{pre}$ is the time when a pre-synaptic neuron fire and $t_{post}$ is the time when a post-synaptic neuron fire. Our system uses equation \ref{eq:LTP} to increase synaptic weights when LTP occurs and uses equation \ref{eq:LTD} to decrease synaptic weights when LTD occurs. 

\subsection{Fault modeling}
We choose neurons and synapses for injecting faults. We create fault excitatory neurons because excitatory neurons receive spikes most often in our simulator and they are prone to breakdown. We create SA0 and SA1 fault synapses between input and excitatory neurons because synaptic weights between them change most frequently and they can easily stuck at specific values. To make fault synapses, we fix synaptic weights to 0 (SA0) or maximum value (SA1).

\begin{figure}[htb!]
\centering
\includegraphics[width=190pt]{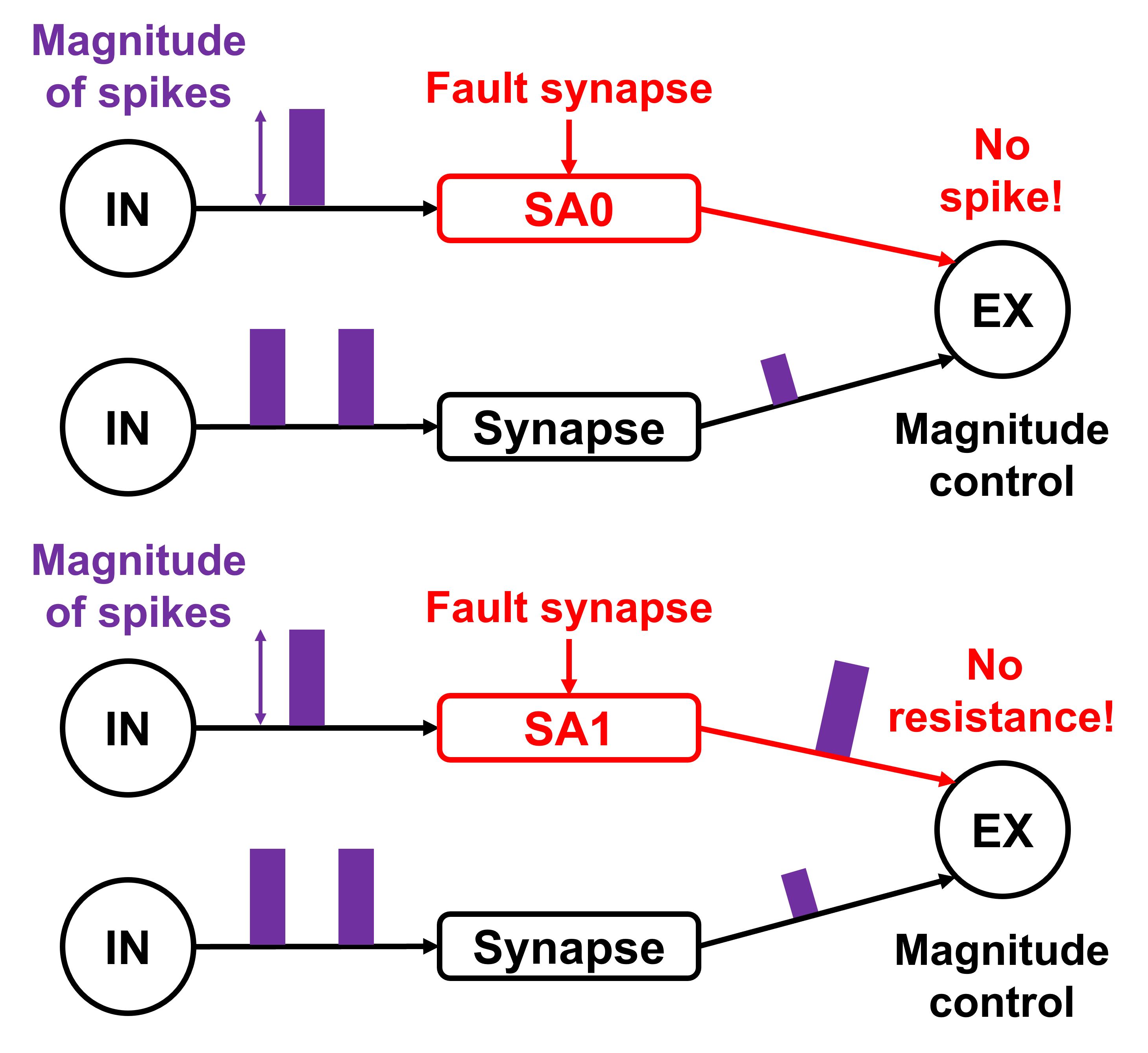}
\caption{SA0 fault synapses block spikes from pre-synaptic neurons and SA1 fault synapses pass spikes from pre-synaptic neurons without any resistance. Therefore, excitatory neurons with fault synapses cannot leran data features appropriately.}
\label{fig:synapses}
\end{figure} 

Figure \ref{fig:synapses} shows how fault synapses affect neuromorphic systems. If SA0 faults occur in a synapse, the synapse does not pass pre-synaptic spikes to a post-synaptic neuron at all. If SA1 faults occur in a synapse, the synapse passes pre-synaptic spikes to the neuron without any resistance. These problems prevent an excitatory neuron with fault synapses from storing data features properly. When a fault occur in excitatory neurons, the fault prevents neurons from increasing their membrane potentials. Therefore, the targeted neurons cannot generate output spikes and become completely inoperable. 

\section{Experiments}
\subsection{Experimental settings}
We use Breast Cancer (BC) and Wine Quality (WQ) dataset to evaluate performance of neuromorphic systems. These datasets are widely used one-dimensional datasets. BC dataset consists of tissue information samples of benign and malignant tumors. Each sample in BC dataset has 30 columns. WQ dataset consists of three kinds of information samples representing wine qualities. Each sample in WQ dataset has 13 columns.

Table \ref{tab:parameters} shows parameter settings we use. We create our neuromorphic system simulator based on BindsNET that is widely used to create neuromorphic system \cite{bindsnet}. To apply memristive characteristics to our system, we adopt non-linearity to weight updates \cite{memristive}. $v$ factors control non-linearity in synaptic weight updates. Non-linearity makes synaptic weights change by a variant value. As $v_{LTP}$ increases, synaptic weights increase more non-linearly when LTP occurs. As $v_{LTD}$ increases, synaptic weights decrease more non-linearly when LTD occurs. If $v$ factors are 0, synaptic weights change by a constant value (linear weight update). We define the two cases in minimum and maximum values of weights. We set $w_{min}$ to 0 and $w_{max}$ to 1 in all excitatory neurons (static G case). We set $w_{min}$ to a random value between 0$\sim$0.5 and $w_{max}$ to a random value between 0.5$\sim$1 in excitatory neurons respectively (random G case). This is because actual synapses elements have slightly different maximum and minimum conductance respectively.

\begin{table}[htb!]
\centering
\begin{tabular}{|l|l|l|}
\hline
\begin{tabular}[c]{@{}l@{}}\textbf{Parameters} \end{tabular} & \textbf{Settings} \\ \hline
\begin{tabular}[c]{@{}l@{}}$v_{LTP}$ \end{tabular}  & -3$\sim$3      \\ \hline
\begin{tabular}[c]{@{}l@{}}$v_{LTD}$ \end{tabular}  & -3$\sim$3      \\ \hline
\begin{tabular}[c]{@{}l@{}}$w_{max}$ \end{tabular}  & 0.5$\sim$1    \\ \hline
\begin{tabular}[c]{@{}l@{}}$w_{min}$ \end{tabular}  & 0$\sim$0.5    \\ \hline
\begin{tabular}[c]{@{}l@{}}$\beta$ \end{tabular}    & 1      \\ \hline
\begin{tabular}[c]{@{}l@{}}$\theta+$ \end{tabular}  & 0.01   \\ \hline
\begin{tabular}[c]{@{}l@{}}$\Delta$t \end{tabular}  & 50ms   \\ \hline
\begin{tabular}[c]{@{}l@{}}Intensity \end{tabular}  & 500    \\ \hline
\begin{tabular}[c]{@{}l@{}}Input neurons \end{tabular}  & 13, 30    \\ \hline
\begin{tabular}[c]{@{}l@{}}Excitatory neurons \end{tabular}  & 10    \\ \hline
\begin{tabular}[c]{@{}l@{}}Inhibitory neurons \end{tabular}  & 10    \\ \hline
\end{tabular}
\caption{Parameter settings for neuromorphic system simulations.}
\label{tab:parameters}
\end{table}

We use accuracy to check the classification performance of neuromorphic systems. We conduct experiments with the following scenarios. First, we measure the accuracy changing memristive characteristics ($v_{LTP}$, $v_{LTD}$, $w_{max}$, and $w_{min}$) without any fault elements. Second, we measure the accuracy changing the five factors: memristive characteristics, fault elements (neurons and synapses), fault ratios, fault synapse types (SA0 and SA1), and fault synapse positions.

\subsection{Analysis on the cases with fault neurons}
\begin{figure}[htb!]
\centering
\subfigure[Accuracy according to the fault neuron ratio with (0, 0) non-linearity factors.]{\includegraphics[width=225pt]{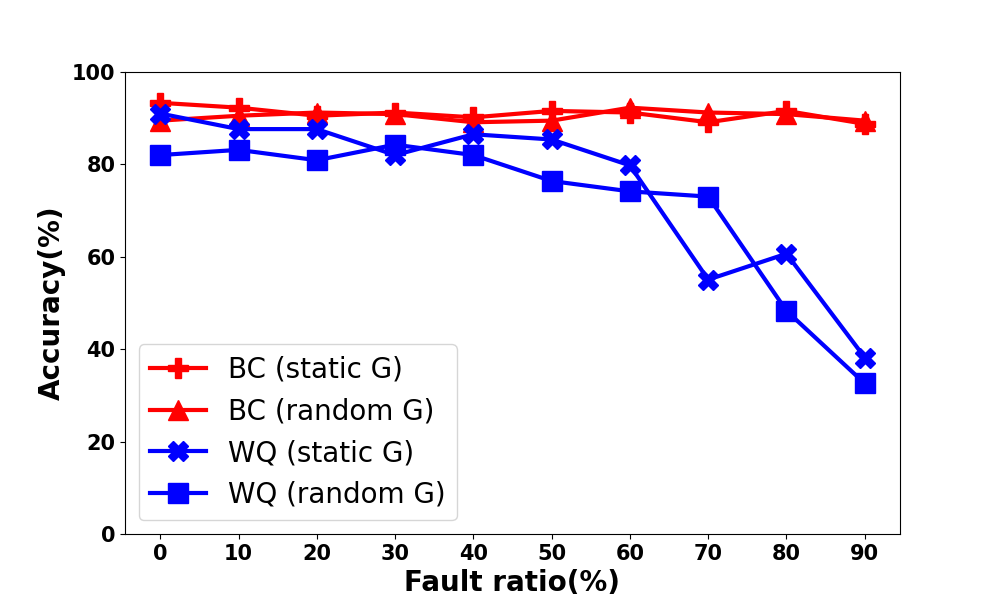}}
\subfigure[Accuracy according the to fault neuron ratio with (3, 3) non-linearity factors.]{\includegraphics[width=225pt]{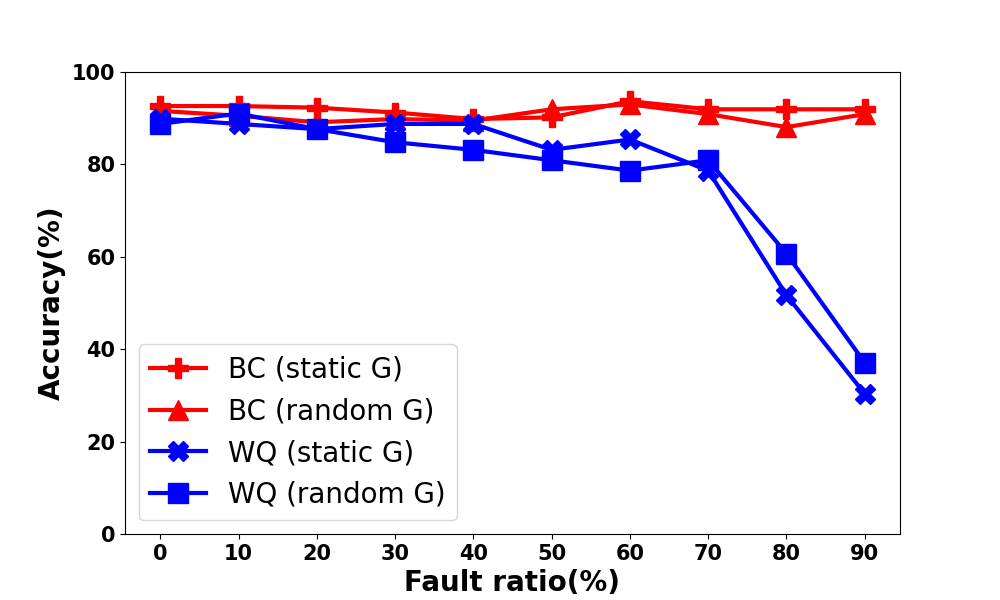}}
\subfigure[Accuracy according to the fault neuron ratio with (-3, -3) non-linearity factors.]{\includegraphics[width=225pt]{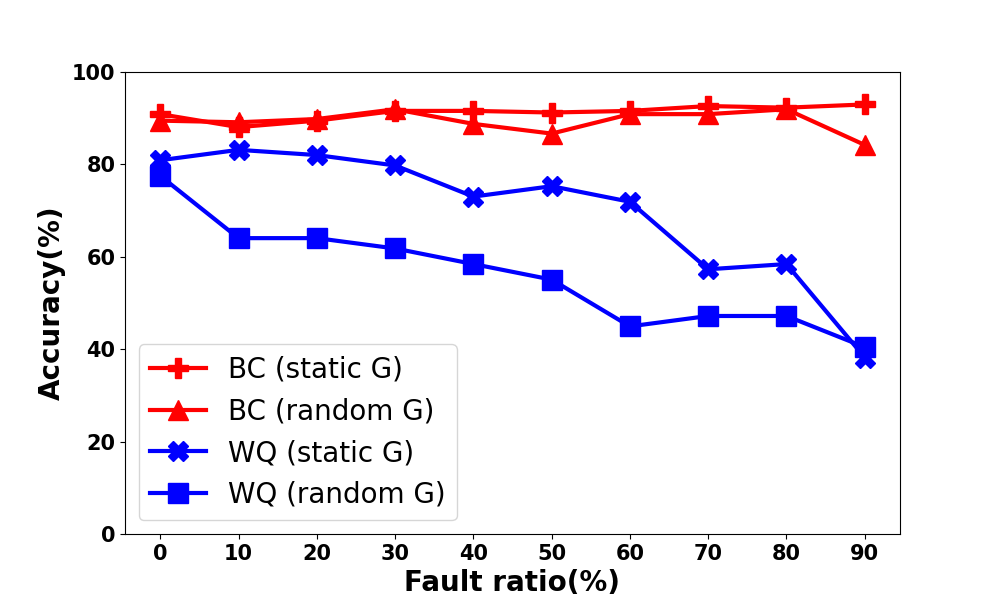}}
\caption{These graphs show accuracy changes caused by fault neurons in different non-linearity factors and G cases.}
\label{fig:FN}
\end{figure}

We change fault the neuron ratio from 0\% to 90\%. We set non-linearity factors ($v_{LTP}$, $v_{LTD}$) to (0, 0), (3, 3), and (-3, -3) \cite{memristive} \cite{concise}. Figure \ref{fig:FN} shows accuracy changes according to fault neurons with (0, 0), (3, 3), (-3, -3) non-linearity factors. With BC dataset, our system maintains its performance although the fault rate increases. However, its performance declines according to the ratio of fault neurons when we use WQ dataset. This is because BC dataset has easy samples to distinguish and consists of only two data classes. Therefore, the system can classify the samples with the small number of operating neurons. On the other hand, WQ dataset has samples that are difficult to distinguish and consists of three data classes. This point makes the system require more than two operating neurons. Therefore, the accuracy drops severely when the number of fault neurons is larger than 7 (the fault ratio is larger than 70\%.). 

When non-linearity factors are (3, 3), the performance change shows similar pattern with (0, 0). This is because (3, 3) makes the system update synaptic weights like (0, 0) \cite{nonlinear}. We find the performance gap between static and random G cases is small with (3, 3). It is derived from the point that the system maintains its stability although $w_{min}$ and $w_{max}$ are random. However, when non-linear factors are (-3, -3), performance is the worst and performance gap is largest among the non-linearity factor settings. This is because (-3, -3) prevents excitatory neurons from representing data features properly due to unstable weight convergence \cite{nonlinear}. With these non-linearity factors, synaptic weights easily converge to specific values (minimum or maximum) and they stuck at them. Therefore, synaptic weights cannot represent features well and performance declines.

\subsection{Analysis on the cases with fault synapses}
To see effects of various fault synapses, we use two types of fault synapses: SA0 and SA1. We make fault synapses in the following two positions for demonstrating the importance of fault synapse positions. First, we inject fault synapses to random positions. Second, we inject fault synapse to important positions corresponding to where significant features are in. We inject them to all excitatory neurons with the same ratios.

\subsubsection{Random position}
\begin{figure}[htb!]
\centering
\subfigure[Accuracy according to the SA0 fault synapse ratio with static and random G cases.]{\includegraphics[width=225pt]{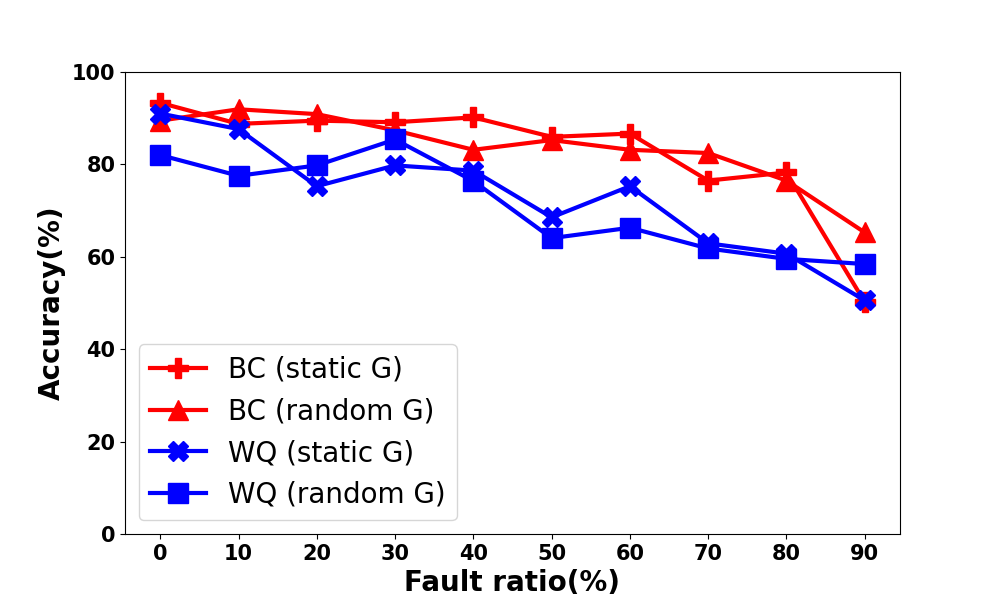}}
\subfigure[Accuracy according to the SA1 fault synapse ratio with static and random G cases.]{\includegraphics[width=225pt]{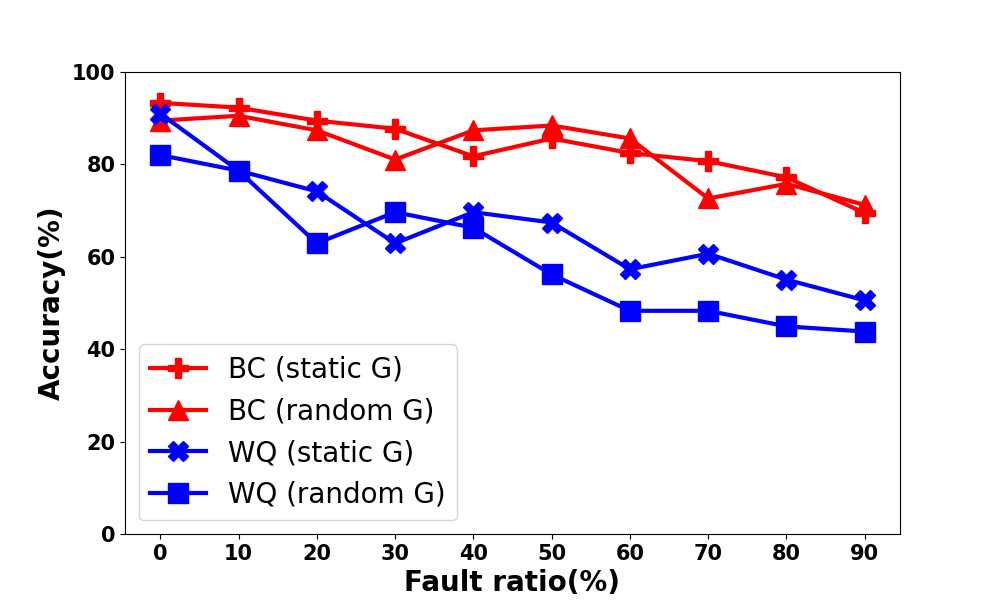}}
\caption{These graphs show accuracy changes caused by fault synapses in random positions with (0, 0) non-linearity factors.}
\label{fig:RFS}
\end{figure}

We inject SA0 and SA1 fault synapses in random positions and compare their effects. Figure \ref{fig:RFS} shows accuracy changes according to the fault synapse ratio with (0, 0) non-linearity factors. Different from fault neurons, the accuracy does not decline significantly. This is because all excitatory neurons can store data features with properly working synapses. We find the accuracy increases although fault ratios increase in some cases. The fault synapses of these cases appear in the position representing unimportant features. Therefore, the system classifies data samples easily in these cases. 

As we can see in figure \ref{fig:RFS}, SA1 fault synapses cause larger performance declines when they appear in random positions. This is because SA1 fault synapses damage the system more severely than SA0 fault synapses due to suppression mechanism. When SA1 fault synapses occur in all excitatory neurons, neurons fire more often because SA1 fault synapses have maximum weight value. Therefore, inhibitory neurons suppress excitatory neurons more than necessary. Then, the decoder cannot choose the marker of a neuron with proper synaptic weights and the answer is wrong in most cases. 

\begin{table}[htb!]
\centering
\begin{tabular}{|c|c|c|c|}
\hline
 & \begin{tabular}[c]{@{}c@{}}(0, 0)\end{tabular} & \begin{tabular}[c]{@{}c@{}}(3, 3)\end{tabular} & \begin{tabular}[c]{@{}c@{}}(-3, -3)\end{tabular} \\ \hline
BC (static G) & 82.84\% & 84\%    & 84.6\%   \\ \hline
BC (random G) & 83.54\% & 85.93\% & 82.6\%   \\ \hline
WQ (static G) & 73.03\% & 73.82\% & 72.81\%  \\ \hline
WQ (random G) & 71.12\% & 73.37\% & 66.63\%  \\ \hline
\end{tabular}
\caption{Average accuracy comparison according to non-linearity factors with SA0 fault synapses in random positions.}
\label{tab:randomSA0}
\end{table}

\begin{table}[htb!]
\centering
\begin{tabular}{|c|c|c|c|}
\hline
 & \begin{tabular}[c]{@{}c@{}}(0, 0)\end{tabular} & \begin{tabular}[c]{@{}c@{}}(3, 3)\end{tabular} & \begin{tabular}[c]{@{}c@{}}(-3, -3)\end{tabular} \\ \hline
BC (static G) & 84\% & 81.83\%    & 83.09\%   \\ \hline
BC (random G) & 82.95\% & 86.77\% & 83.75\%   \\ \hline
WQ (static G) & 66.74\% & 64.95\% & 70\%  \\ \hline
WQ (random G) & 60.11\% & 66.73\% & 58.91\%  \\ \hline
\end{tabular}
\caption{Average accuracy comparison according to non-linearity factors with SA1 fault synapses in random positions.}
\label{tab:randomSA1}
\end{table}

Table \ref{tab:randomSA0} and \ref{tab:randomSA1} show performance comparison according to non-linear factors with fault synapses in random positions. In particular, there are no significant differences between (0, 0) and (3, 3) non-linearity factors. However, (-3, -3) shows the worst performance and largest performance gap between G cases among the non-linearity factors. This is because synaptic weights without faults converge abnormally due to (-3, -3). Therefore, normal synapses cannot represent data features properly even though they do not have faults.

\subsubsection{Important position}
\begin{figure}[htb!]
\centering
\includegraphics[width=225pt]{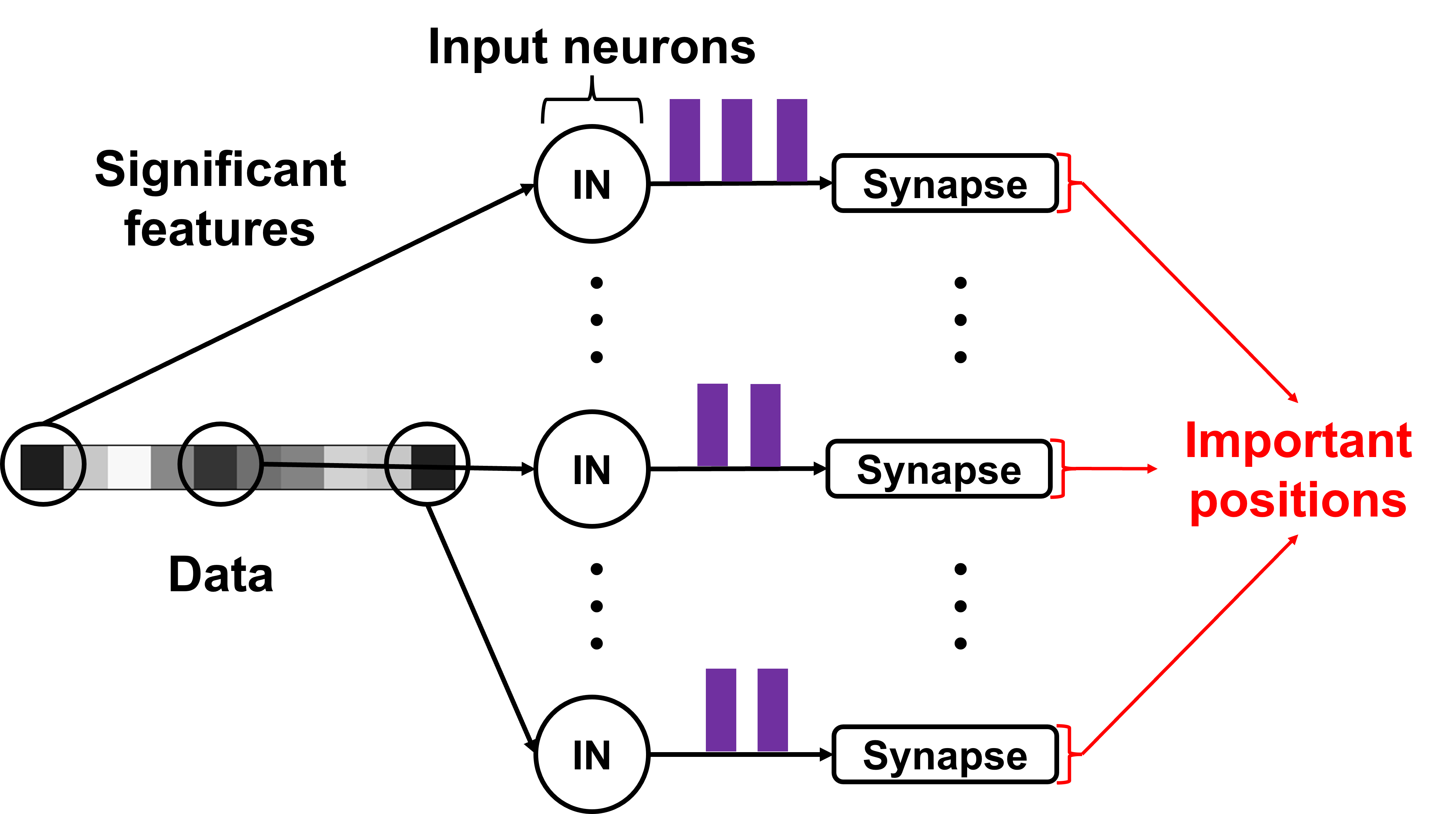}
\caption{When fault synapses appear in important positions, the spikes caused by significant features in input data are blocked or passed to excitatory neurons according to fault types.}
\label{fig:position}
\end{figure}

In figure \ref{fig:position}, black squares are significant features having large values and white squares are unimportant features having negligibly small values. The black squares are important to classify samples. Synapses corresponding to the black squares are in important positions and they enter spikes from significant features of data. On the other hand, the white and light gray squares are unimportant. Synapses corresponding to them are in unimportant positions and they enter spikes from less important features. If fault synapses occur in the important positions, the faults prevent excitatory neurons from storing proper features of data severely.

\begin{figure}[htb!]
\centering
\subfigure[Accuracy according to the SA0 fault synapse ratio with static and random G cases.]{\includegraphics[width=225pt]{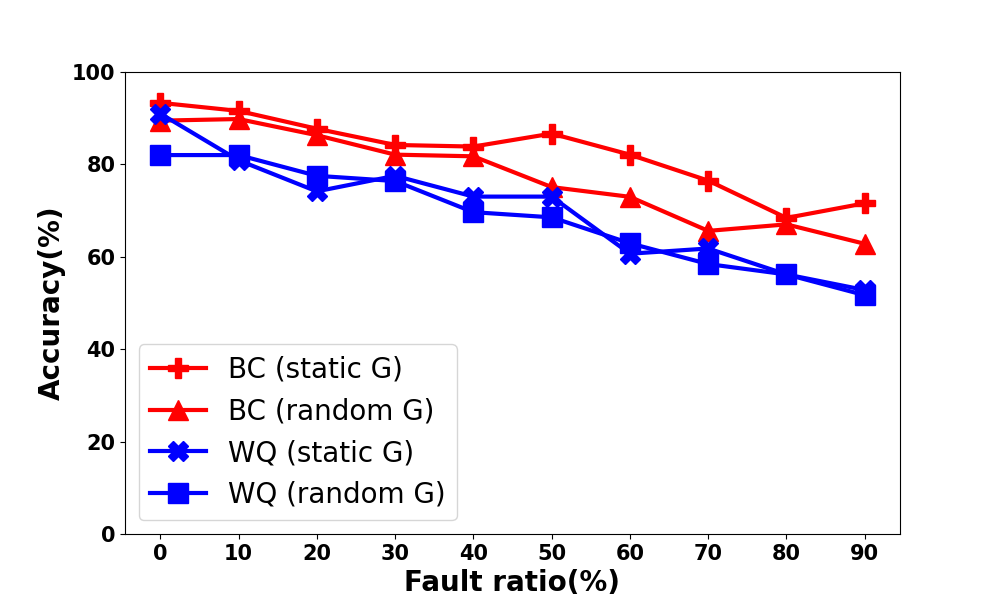}}
\subfigure[Accuracy according to the SA1 fault synapse ratio with static and random G cases.]{\includegraphics[width=225pt]{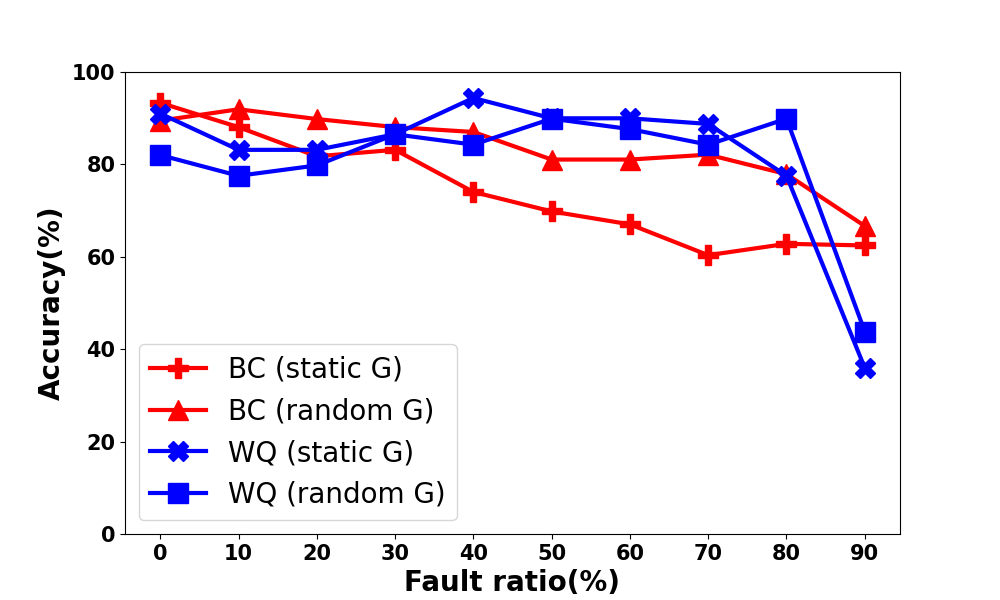}}
\caption{These graphs show accuracy changes caused by fault synapses in important positions with (0, 0) non-linearity factors.}
\label{fig:IFS}
\end{figure}

We inject SA0 and SA1 fault synapses in important position and analyze how performance changes. Figure \ref{fig:IFS} shows accuracy changes according to the fault synapse ratio when non-linearity factors are (0, 0). It is interesting that accuracy increases although SA1 fault synapses appear in important positions. This is because SA1 fault synapses pass all input spikes from significant features when they appear in important features. Therefore, excitatory neurons can learn significant features easily and the system classifies samples clearly. 

When SA0 fault synapses appear in important positions, accuracy declines steadily. This is because fault synapses appear in important positions. If SA0 fault synapses appear in positions corresponding to unimportant features, they help the system to classify samples well since the they prevent unimportant features from passing to excitatory neurons. However, inject them to only important positions and they cannot appear in unimportant positions. Therefore, the performance keeps decreasing in this case.  

\begin{table}[htb!]
\centering
\begin{tabular}{|c|c|c|c|}
\hline
 & \begin{tabular}[c]{@{}c@{}}(0, 0)\end{tabular} & \begin{tabular}[c]{@{}c@{}}(3, 3)\end{tabular} & \begin{tabular}[c]{@{}c@{}}(-3, -3)\end{tabular} \\ \hline
BC (static G) & 82.6\% & 83.12\%    & 83.51\%   \\ \hline
BC (random G) & 83.54\% & 79.33\% & 78.77\%   \\ \hline
WQ (static G) & 73.03\% & 72.92\% & 66.29\%  \\ \hline
WQ (random G) & 71.12\% & 63.6\% & 62.47\%  \\ \hline
\end{tabular}
\caption{Average accuracy comparison according to non-linearity factors with SA0 fault synapses in important positions.}
\label{tab:importantSA0}
\end{table}

\begin{table}[htb!]
\centering
\begin{tabular}{|c|c|c|c|}
\hline
& \begin{tabular}[c]{@{}c@{}}(0, 0)\end{tabular} & \begin{tabular}[c]{@{}c@{}}(3, 3)\end{tabular} & \begin{tabular}[c]{@{}c@{}}(-3, -3)\end{tabular} \\ \hline
BC (static G) & 74.28\% & 76.21\%    & 75.41\%   \\ \hline
BC (random G) & 83.51\% & 81.35\% & 78.88\%   \\ \hline
WQ (static G) & 82.04\% & 82.45\% & 78.77\%  \\ \hline
WQ (random G) & 80.56\% & 82.13\% & 75.39\%  \\ \hline
\end{tabular}
\caption{Average accuracy comparison according to non-linearity factors with SA1 fault synapses in important positions.}
\label{tab:importantSA1}
\end{table}

Table \ref{tab:importantSA0} and \ref{tab:importantSA1} show performance comparison according to non-linearity factors with fault synapses in important positions. When non-linearity factors are (3, 3), accuracy is similar to that of (0, 0) non-linearity factors. However, accuracy is the lowest when non-linearity factors are (-3, -3). This is because (-3, -3) makes the synaptic weights change by large values. Since synapses in important positions cannot be updated, synapses without faults represent unimportant features. However, unimportant features are not noticeable and the synaptic weights should change minutely to represent the features. (-3, -3) prevents them from changing minutely and this problem reduces the performance.
\section{Conclusion}
In this paper, we investigated effects of fault elements on neuromorphic systems in various scenarios for the first time. Our analysis shows how the performance changes with fault elements with various fault ratios, fault positions, and memristive characteristics. We discovered the following four important points. First, non-linear weight updates by memristors make neuromorphic systems vulnerable to fault elements in some cases. Second, there are fault neuron ratios making the performance drop sharply. Third, SA1 fault synapses damage systems more severely than SA0 fault synapses when they appear in random positions. Finally, SA1 fault synapses increase the performance and SA0 fault synapses reduce the performance when they appear in important positions. Our analysis gives guidelines to deal with fault elements in actual neuromorphic systems. Our ongoing works will be proposing an adaptive fault recovery system according to fault ratios, types, and positions in actual neuromorphic systems. 
\section*{Acknowledgement}
This work was sponsored by the National Research Foundation of Korea (NRF) (grant no. NRF-2021M3F3A2A01037962).

\end{document}